\documentclass{article} 
\usepackage[accepted]{icml2017}
\usepackage{times}
\usepackage{url}
\usepackage{graphicx}
\usepackage{multirow,tabularx}
\usepackage{amsmath}
\usepackage{wrapfig}
\usepackage{xspace}

\usepackage{bm}
\usepackage{bbm}
\usepackage{listings}
\usepackage{multirow}
\usepackage{booktabs}
\usepackage{subcaption} 
\usepackage{color}

\newcommand{\figref}[1]{Fig.~\ref{#1}}

\newcommand{\terpret}{{\sc TerpreT}\xspace}
\newcommand{\neuralterpret}{{\sc Neural TerpreT}\xspace}

\newcommand{\neuralterpretshort}{NTPT\xspace}
\newcommand{\secref}[1]{Sec.~\ref{#1}}
\newcommand{\ourproblem}{Lifelong Perceptual Programming By Example\xspace}
\newcommand{\ourproblemshort}{LPPBE\xspace}
\newcommand{\rev}[1]{#1}

\newcommand{\sumsScenario}{\textsc{Add2x2}\xspace}
\newcommand{\arithScenario}{\textsc{Apply2x2}\xspace}
\newcommand{\mathScenario}{\textsc{Math}\xspace}

\DeclareFixedFont{\ttb}{T1}{txtt}{bx}{n}{8} 
\DeclareFixedFont{\ttm}{T1}{txtt}{m}{n}{8}  
\definecolor{deepblue}{rgb}{0,0,0.5}
\definecolor{deepred}{rgb}{0.6,0,0}
\definecolor{deepgreen}{rgb}{0,0.5,0}
\definecolor{gray}{rgb}{0.3,0.3,0.3}
\lstnewenvironment{python}[1][]
{
\pythonstyle
\lstset{#1}
}
{}

\newcommand\pythonstyle{\lstset{
language=Python,
basicstyle=\tiny\ttm,
moredelim=**[is][\color{gray}]{~}{~},
moredelim=**[is][\tiny\ttb\color{deepblue}]{!}{!},
otherkeywords={self, with, pass, as},             
keywordstyle=\tiny\ttb\color{deepblue},
numbers=none,
stepnumber=5,
firstnumber=1,
numberfirstline=false,
numberstyle=\tiny,
emph={@Runtime, @Learn, Param, Var},          
emphstyle=\ttb\color{deepred},    
commentstyle=\color{deepgreen},
frame=tb,                         
showstringspaces=false            %
}}

\icmltitlerunning{Differentiable Programs with Neural Libraries}

\begin{document}
\twocolumn[
\icmltitle{\rev{Differentiable Programs with Neural Libraries}}




\begin{icmlauthorlist}
\icmlauthor{Alexander L. Gaunt}{MSR}
\icmlauthor{Marc Brockschmidt}{MSR}
\icmlauthor{Nate Kushman}{MSR}
\icmlauthor{Daniel Tarlow}{Google}
\end{icmlauthorlist}

\icmlaffiliation{MSR}{Microsoft Research, Cambridge, UK}
\icmlaffiliation{Google}{Google Brain, Montr\'{e}al, Canada (work done while at Microsoft)}

\icmlcorrespondingauthor{Alexander L. Gaunt}{algaunt@microsoft.com}

\icmlkeywords{program induction}

\vskip 0.3in
]

\printAffiliationsAndNotice{}

\begin{abstract}
\rev{
We develop a framework for combining differentiable programming languages with neural networks. Using this framework we create end-to-end trainable systems that learn to write interpretable algorithms with perceptual components. We explore the benefits of inductive biases for strong generalization and modularity that come from the program-like structure of our models. In particular, modularity allows us to learn a library of (neural) functions which grows and improves as more tasks are solved. Empirically, we show that this leads to lifelong learning systems that transfer knowledge to new tasks more effectively than baselines.
}
\end{abstract}

\comment{
introduce and develop solutions for the problem of \emph{\ourproblem~(\ourproblemshort)}.
  The problem is to induce a series of programs that require understanding perceptual data like images or text.
  \ourproblemshort~systems learn from weak supervision
  (input-output examples) and incrementally construct a shared library of components that grows and improves
  as more tasks are solved.
  Methodologically, we extend differentiable interpreters to operate on perceptual data and to share components
  across tasks. Empirically we show that this leads to a lifelong learning system that transfers
  knowledge to new tasks more effectively than baselines, and the performance on earlier tasks continues
  to improve even as the system learns on new, different tasks.
\end{abstract}
}

\section{Introduction}
\begin{figure*}[t]
\centering
\includegraphics[width=\textwidth]{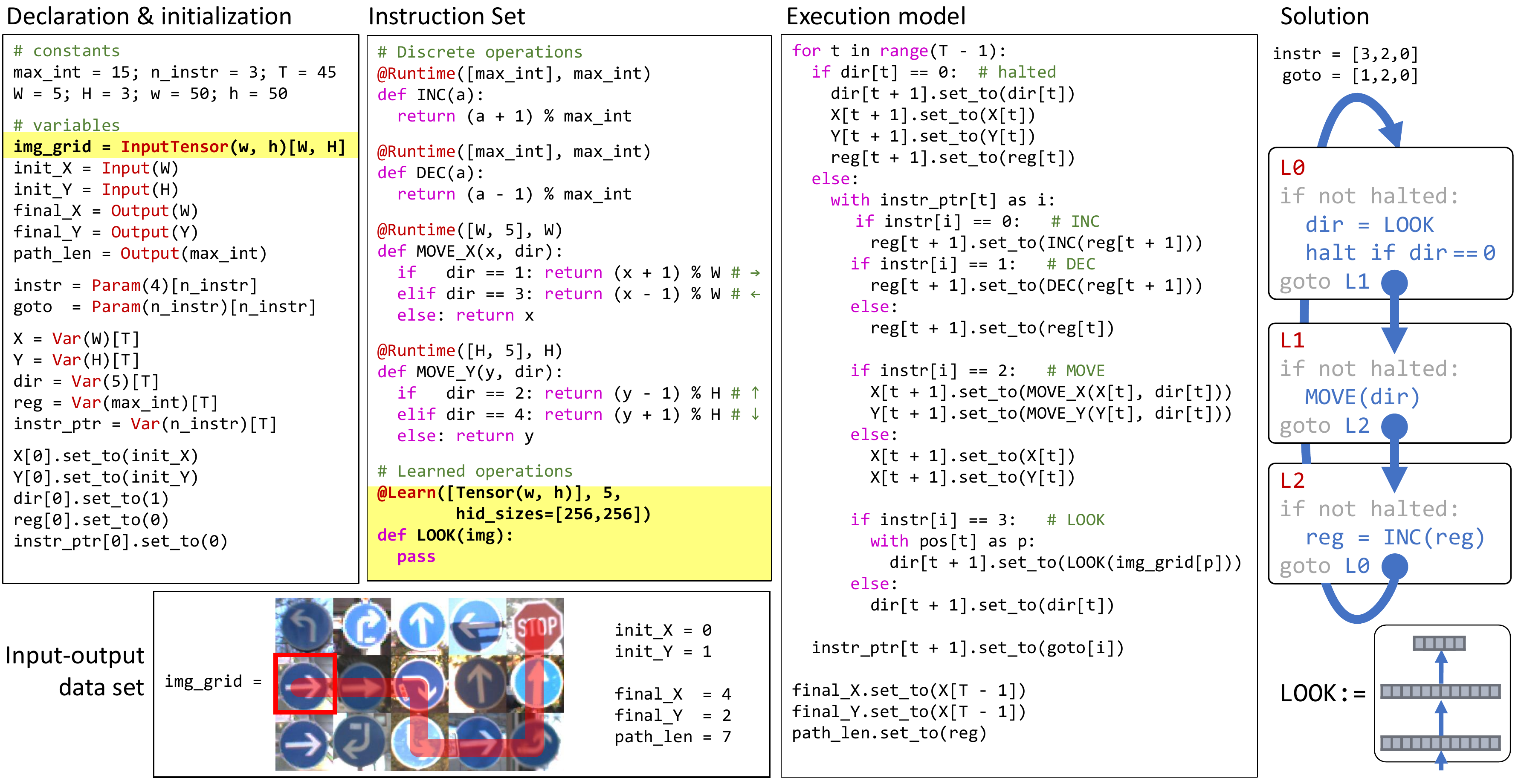}
\caption{\small Components of an illustrative \neuralterpretshort program for learning loopy programs that measure path length (\texttt{path\_len}) through a maze of street sign images. The learned program (parameterized by \texttt{instr} and \texttt{goto}) must control the position (\texttt{X}, \texttt{Y}) of an agent on a grid of (\texttt{W}$\times$\texttt{H}) street sign images each of size (\texttt{w}$\times$\texttt{h}). The agent has a single register of memory (\texttt{reg}) and learns to interpret street signs using the \texttt{LOOK} neural function. A solution consists of a correctly inferred program and a trained neural network. Learnable components are shown in blue and the \neuralterpretshort extensions to the \terpret language are highlighted. The red path on the \texttt{img\_grid} shows the desired behavior and is not provided at training time.}
\label{fig:navigation}
\vspace{-2ex}
\end{figure*}

\rev{
Recently, there has been much work on learning algorithms using neural networks.
Following the idea of the Neural Turing Machine~\cite{Graves14}, this work has
focused on extending neural networks with interpretable components that are
differentiable versions of traditional computer components, such as external
memories, stacks, and discrete functional units.
However, trained models are not easily interpreted as the learned algorithms are
embedded in the weights of a monolithic neural network.
In this work we flip the roles of the neural network and differentiable computer
architecture.
We consider \emph{interpretable} controller architectures which express algorithms
using differentiable programming languages~\cite{Gaunt16,Riedel16,Bunel16}.
In our framework, these controllers can execute discrete functional units (such as
those considered by past work), but also have access to a library of
trainable, uninterpretable neural network functional units.
The system is end-to-end differentiable such that the source code representation
of the algorithm is jointly induced with the parameters of the neural
function library.
In this paper we explore potential advantages of this class of hybrid model over
purely neural systems, with a particular emphasis on lifelong learning systems
that learn from weak supervision.


We concentrate on \emph{perceptual programming by example} (PPBE) tasks that have both algorithmic and perceptual elements to exercise the traditional strengths of program-like and neural components. Examples of this class of task include navigation tasks guided by images or natural language (see \figref{fig:navigation}) or handwritten symbol manipulation (see \secref{sec:tasks}). Using an illustrative set of PPBE tasks we aim to emphasize two specific benefits of our hybrid models:
 
First, the source code representation in the controller allows modularity: the neural components are small functions that specialize to different tasks within the larger program structure. It is easy to separate and share these functional units to transfer knowledge between tasks. In contrast, the absence of well-defined functions in purely neural solutions makes effective knowledge transfer more difficult, leading to problems such as catastrophic forgetting in multitask and lifelong learning \cite{mccloskey89, ratcliff90}.
In our experiments, we consider a lifelong learning setting in which we train the system on a \emph{sequence} of PPBE tasks that share perceptual subtasks.

Second, the source code representation enforces an inductive bias that favors
learning solutions that exhibit strong generalization.
For example, once a suitable control flow structures (e.g., a \texttt{for}
loop) for a list manipulation problem was learned on short examples, it
trivially generalizes to lists of arbitrary length.
In contrast, although some neural architectures demonstrate a surprising
ability to generalize, the reasons for this generalization are not fully
understood~\cite{Zhang17} and generalization performance invariably degrades as
inputs become increasingly distinct from the training data.


This paper is structured as follows. We first present a language, called
\neuralterpret (\neuralterpretshort), for specifying hybrid source code/neural
network models (\secref{sec:methods}), and then introduce a sequence of PPBE
tasks (\secref{sec:tasks}).
Our NTPT models and purely neural baselines are described in \secref{sec:models}
and \ref{sec:baselines} respectively.
The experimental results are presented in \secref{sec:experiments}.



}

\comment{
A goal of artificial intelligence is to build a single large neural network model that can be trained in a \emph{lifelong learning} setting; i.e., on a sequence of diverse tasks over a long period of time, and gain cumulative knowledge about different domains as it is presented with new tasks.
The hope is that such systems will learn more accurately and from less data than existing systems, and that they will exhibit more flexible intelligence.
However, 
despite some work showing promise towards multitask learning (training on many tasks at once) and transfer learning (using source tasks to improve learning in a later target task) \citep{Caruana97,luong2015multi,parisotto2016,rusu16},
most successes of neural networks today come from training a single network on a single task,
indicating that this goal is highly challenging to achieve.

We argue for two properties that such systems should have in addition to the ability to learn from a sequence of diverse tasks.
First is the ability to learn from weak supervision.
Gathering high-quality labeled datasets is expensive, and this effort is multiplied if all tasks require strong labelling.
In this work, we focus on weak supervision in the form of pairs of input-output examples that come from executing simple programs with no labelling of intermediate states.
Second is the ability to distill knowledge into subcomponents that can be shared across tasks.
If we can learn models where the knowledge about shared subcomponents is disentangled from task-specific knowledge, then the sharing of knowledge across tasks will likely be more effective.
Further, by isolating shared subcomponents, we expect that we could develop systems that exhibit reverse transfer (i.e., performance on earlier tasks automatically improves by improving the shared components in later tasks).

A key challenge in achieving these goals with neural models is the difficulty in interpreting weights inside a trained network.
Most notably, with a purely neural model, 
subcomponents of knowledge gained after training on one task cannot be easily transferred to related tasks.
Conversely, traditional computer programs naturally structure solutions to diverse problems in an interpretable, modular form allowing (1) re-use of subroutines in solutions to new tasks and (2) modification or error correction by humans.
Inspired by this fact, we develop end-to-end trainable models that structure their solutions as a library of functions, some of which are represented as source code, and some of which are neural networks.

Methodologically, we start from recent work on programming by example (PBE) with differentiable interpreters, which shows that it is possible to use gradient descent to induce source code operating on basic data types (e.g. integers) from input-output examples \citep{Gaunt16, Riedel16,Bunel16}.
In this work we combine these differentiable interpreters with neural network classifiers in an end-to-end trainable system that learns programs that manipulate perceptual data.
In addition, we make our interpreter modular, which allows \emph{lifelong learning} on a sequence of related tasks: rather than inducing one fresh program per task, the system is able to incrementally build a library of (neural) functions that are shared across task-specific programs.
To encapsulate the challenges embodied in this problem formulation, we name the
problem \emph{\ourproblem~(\ourproblemshort)}.
Our extension of differentiable interpreters that allows perceptual data types, neural network function definitions, and lifelong learning
is called \emph{\neuralterpret~(\neuralterpretshort)}.

Empirically, we show that a \neuralterpretshort-based model 
learns to perform a sequence of tasks based on images of digits and mathematical operators.
In early tasks, the model learns the concepts of digits and mathematical operators from a variety of weak supervision, then in a
later task it learns to compute the results of variable-length mathematical expressions.
The approach is resilient to catastrophic forgetting \citep{mccloskey89, ratcliff90};
on the contrary, results show that performance continues to improve on earlier tasks even when only training on later tasks.
In total, the result is a method that can gather knowledge from a variety of weak supervision, distill it into a cumulative, re-usable library, and use the library within induced algorithms to exhibit strong generalization.
}

\section{Building hybrid models}
\label{sec:methods}

\begin{figure*}
\vspace{-0.3cm}
\centering
\includegraphics[width=\textwidth]{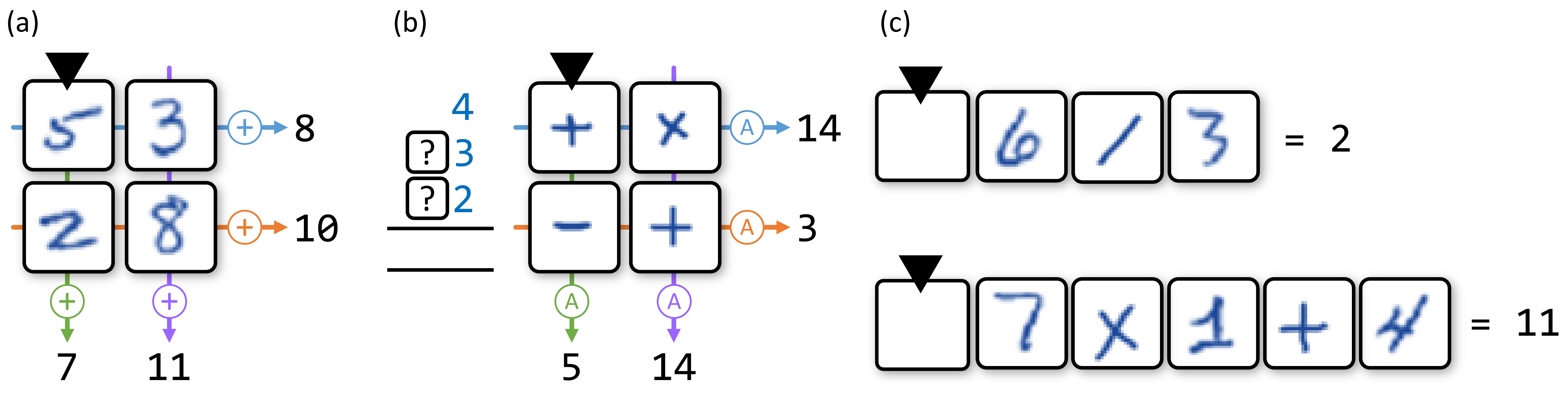}
\caption{\label{fig:tasks} \small Overview of tasks in the (a) \sumsScenario, (b) \arithScenario and (c) \mathScenario scenarios. `A' denotes the \texttt{APPLY} operator which replaces the ? tiles with the selected operators and executes the sum. We show two \mathScenario examples of different length. }
\vspace{-3mm}
\end{figure*}

\rev{
The \terpret~language \citep{Gaunt16} provides a system for constructing differentiable program interpreters that can induce source code operating on basic data types (e.g. integers) from input-output examples. We extend this language with the concept of learnable neural functions. These can either be embedded inside the differentiable interpreter as mappings from integer to integer or (as we emphasize in this work) can act as learnable  interfaces between perceptual data represented as floating point \texttt{Tensor}s and the differentiable interpreter's integer data type. Below we briefly review the \terpret~language and describe the \neuralterpret extensions.
}

\subsection{\terpret}
\terpret~programs specify a differentiable interpreter by defining the
relationship between \texttt{Input}s and \texttt{Output}s via a set of
inferrable \texttt{Param}s (that define an executable program) and \texttt{Var}s
(that store intermediate results).
\terpret~requires all of these variables to range over bounded integers.
The model is made differentiable by a
compilation step that lifts the relationships between integers specified by the
\terpret code to relationships between marginal distributions over integers in
finite ranges.
\figref{fig:navigation} illustrates an example application of the language.

\terpret can be translated into a TensorFlow \citep{abadi16} computation graph which can then be trained using standard methods.
For this, two key features of the language need to be translated:

\begin{itemize}
\item \textbf{Function application}. The statement \texttt{z.set\_to(foo(x, y))} is translated into {$\mu^z_i = \sum_{jk} I_{ijk}\mu^x_j\mu^y_k$} where ${\bm\mu}^a$ represents the marginal distribution for the variable $a$ and ${\bm I}$ is an indicator tensor $\mathbbm{1}[i = \texttt{foo}(j,k)]$. This approach extends to all functions mapping any number of integer arguments to an integer output.
\item \textbf{Conditional statements} The statements \texttt{if x == 0: z.set\_to(a); elif x == 1: z.set\_to(b)} are translated to $\bm\mu^z = \mu^x_0\bm\mu^a+\mu^x_1\bm\mu^b$. More complex statements follow a similar pattern, with details given in \cite{Gaunt16}.
\end{itemize}

\subsection{\neuralterpret}
To handle perceptual data, we relax the restriction that all variables need to be
finite integers.
We introduce a new \texttt{Tensor} type whose dimensions are fixed at declaration,
and which is suitable for storing perceptual data.
Additionally, we introduce \emph{learnable functions} that can process integer or tensor
variables.
A learnable function is declared using
 \texttt{@Learn([$d_1, \ldots, d_D$], $d_{out}$, hid\_sizes=[$\ell_1, \ldots, \ell_L$])},
where the first component specifies the dimensions (resp. ranges) $d_1, \ldots, d_D$ of the
input tensors (resp. integers) and the second specifies the dimension of
the output.
\neuralterpretshort compiles such functions into a fully-connected feed-forward
neural network whose layout is controlled by the \texttt{hid\_sizes} component (specifying the number neurons in each layer).
The inputs of the function are simply concatenated.
Tensor output is generated by learning a mapping from the last hidden layer, and
finite integer output is generated by a softmax layer producing a distribution
over integers up to the declared bound.
Learnable parameters for the generated network are shared across every use of the function in the
\neuralterpretshort program, and as they naturally fit into the computation graph for
the remaining \terpret program, the whole system is trained end-to-end.
\rev{We illustrate an example \neuralterpretshort program for learning navigation tasks in a maze of street signs \citep{Stallkamp11} in
\figref{fig:navigation}.}

\section{A Lifetime of PPBE Tasks}
\label{sec:tasks}

\rev{
Motivated by the hypothesis that the modularity of the source code representation benefits knowledge transfer, we devise a \emph{sequence} of PPBE tasks to be solved by sharing knowledge between tasks. Our tasks are based on algorithmic manipulation of handwritten digits and mathematical operators.

In early tasks the model learns to navigate simple $2\times 2$ grids of images, and to become familiar with the concepts of digits and operators from a variety of weak supervision. Despite their simplicity, these challenges already pose problems for purely neural lifelong learning systems.

The final task in the learning lifetime is more complex and designed to test generalization properties: the system must learn to compute the results of variable-length mathematical expressions expressed using handwritten symbols. The algorithmic component of this task is similar to arithmetic tasks presented to contemporary Neural GPU models~\cite{Kaiser15,Price16}. The complete set of tasks is illustrated in \figref{fig:tasks} and described in detail below.
}

\paragraph{\sumsScenario scenario:}
The first scenario in \figref{fig:tasks}(a) uses of a $2 \times 2$ grid of MNIST
digits.
We set 4 tasks based on this grid: compute the sum of the digits in the (1) top
row, (2) left column, (3) bottom row, (4) right column.
All tasks require classification of MNIST digits, but need different programs to
compute the result.
As training examples, we supply \emph{only} a grid and the resulting sum.
Thus, we \emph{never} directly label an MNIST digit with its class.

\paragraph{\arithScenario scenario:}
The second scenario in \figref{fig:tasks}(b) presents a $2 \times 2$
grid of of handwritten arithmetic operators.
Providing three auxiliary random integers 
 \texttt{d}$_1$, \texttt{d}$_2$, \texttt{d}$_3$,
we again set 4 tasks based on this grid, namely to evaluate the expression\footnote{Note that for simplicity, our toy system ignores operator precedence and executes operations from left to right - i.e. the sequence in the text is executed as \texttt{((d$_1$ op$_1$ d$_2$) op$_2$ d$_3$)}.}
\texttt{d$_1$ op$_1$ d$_2$ op$_2$ d$_3$} where (\texttt{op$_1$},
\texttt{op$_2$}) are the operators represented in the (1) top row, (2) left
column, (3) bottom row, (4) right column.
In comparison to the first scenario, the dataset of operators is relatively
small and consistent\footnote{200 handwritten examples of each operator were
  collected from a single author to produce a training set of 600 symbols and a
  test set of 200 symbols from which to construct random $2\times 2$ grids.},
making the perceptual task of classifying operators considerably
easier.
However, the algorithmic part is more difficult, requiring non-linear operations
on the supplied integers.

\paragraph{\mathScenario scenario:}
The final task in \figref{fig:tasks}(c) requires combination of the knowledge gained
from the weakly labeled data in the first two scenarios to execute a handwritten
arithmetic expression.

\section{Models}
\label{sec:models}

\begin{figure}
\centering
\begin{subfigure}[t]{0.2\textwidth}
\begin{minipage}[t]{0.9\textwidth}
(a)
\begin{python}
# initialization:
~R0 = READ~
# program:
~R1 =~ MOVE_EAST
~R2 =~ MOVE_SOUTH
~R3 =~ SUM(R0, R1)
~R4 =~ NOOP
return R3
\end{python}
\end{minipage}
\end{subfigure}
\begin{subfigure}[t]{.2\textwidth}
\begin{minipage}[t]{1.1\textwidth}
(b)
\begin{python}
# initialization:
~R0 = InputInt[0]~
~R1 = InputInt[1]~
~R2 = InputInt[2]~
~R3 = READ~
# program:
~R4 =~ MOVE_EAST
~R5 =~ MOVE_SOUTH
~R6 =~ APPLY(R0, R1, R4)
~R7 =~ APPLY(R6, R2, R5)
return R7
\end{python}
\end{minipage}
\end{subfigure}
\caption{\small Example solutions for the tasks on the right columns of the (a) \sumsScenario and (b) \arithScenario scenarios. The read head is initialized \texttt{READ}ing the top left cell and any auxiliary \texttt{InputInt}s are loaded into memory. Instructions and arguments shown in black must be learned\label{fig:solutions}.}
\vspace{-0.5cm}
\end{figure}

\rev{
We study two kinds of \neuralterpretshort model. First, for navigating the introductory $2\times 2$ grid scenarios, we create a model which learns to write simple straight-line code. Second, for the \mathScenario scenario we ask the system to use a more complex language which supports loopy control flow  (note that the baselines will also be specialized between the $2\times 2$ scenarios and the \mathScenario scenario). Knowledge transfer is achieved by defining a library of 2 neural network functions shared across all tasks and scenarios. Training on each task should produce a task-specific source code solution (from scratch) and improve the overall usefulness of the shared networks. Below we outline further details of the models.
}

\subsection{Shared components}
\label{sec:shared}
We refer to the 2 networks in the shared library as \texttt{net\_0} and \texttt{net\_1}. Both networks have similar architectures: they take a $28\times 28$ monochrome image as input and pass this sequentially through two fully connected layers each with 256 neurons and ReLU activations. The last hidden vector is passed through a fully connected layer and a softmax to produce a 10 dimensional output (\texttt{net\_0}) or 4 dimensional output (\texttt{net\_1}) to feed to the differentiable interpreter (the output sizes are chosen to match the number of classes of MNIST digits and arithmetic operators respectively).

One restriction that we impose is that when a new task is presented, no more than one new untrained network can be introduced into the library (i.e. in our experiments the very first task has access to only \texttt{net\_0}, and all other tasks have access to both nets). This restriction is imposed because if a differentiable program tries to make a call to one of $N$ untrained networks based on an unknown parameter \texttt{net\_choice = Param(N)}, then the system effectively sees the $N$ nets together with the \texttt{net\_choice} parameter as one large untrained network, which cannot usefully be split apart into the $N$ components after training.

\subsection{$2 \times 2$ model}
For the $2 \times 2$ scenarios we build a model capable of writing short straight line algorithms with up to 4 instructions. The model consists of a read head containing \texttt{net\_0} and \texttt{net\_1} which are connected to a set of registers each capable of holding integers in the range $0,\ldots, M$, where $M=18$. The head is initialized reading the top left cell of the $2\times 2$ grid. At each step in the program, one instruction can be executed, and lines of code are constructed by choosing an instruction and addresses of arguments for that instruction. We follow \cite{Feser16} and allow each line to store its result in a separate immutable register. For the \sumsScenario scenario the instruction set is:

\begin{itemize}
\item \texttt{NOOP}: a trivial no-operation instruction.
\item \texttt{MOVE\_NORTH, MOVE\_EAST, MOVE\_SOUTH, MOVE\_WEST}: translate the head (if possible) and return the result of applying the neural network chosen by \texttt{net\_choice} to the image in the new cell.
\item \texttt{ADD($\cdot, \cdot$)}: accepts two register addresses and returns the sum of their contents.
\end{itemize}

The parameter \texttt{net\_choice} is to be learned and decides which of \texttt{net\_0} and \texttt{net\_1} to apply. In the \arithScenario scenario we extend the \texttt{ADD} instruction to \texttt{APPLY(a, b, op)} which interprets the integer stored at \texttt{op} as an arithmetic operator and computes\footnote{All operations are performed modulo $(M+1)$ and division by zero returns $M$.} \texttt{a op b}. In addition, for the \arithScenario scenario we initialize three registers with the auxiliary integers supplied with each $2 \times 2$ operator grid [see \figref{fig:tasks}(b)]. In total, this model exposes a program space of up to $\sim 10^{12}$ syntactically distinct programs.

\subsection{\mathScenario model}
\begin{figure}
\centering
\begin{subfigure}[t]{.28\textwidth}
(a) \\
\includegraphics[width=2in]{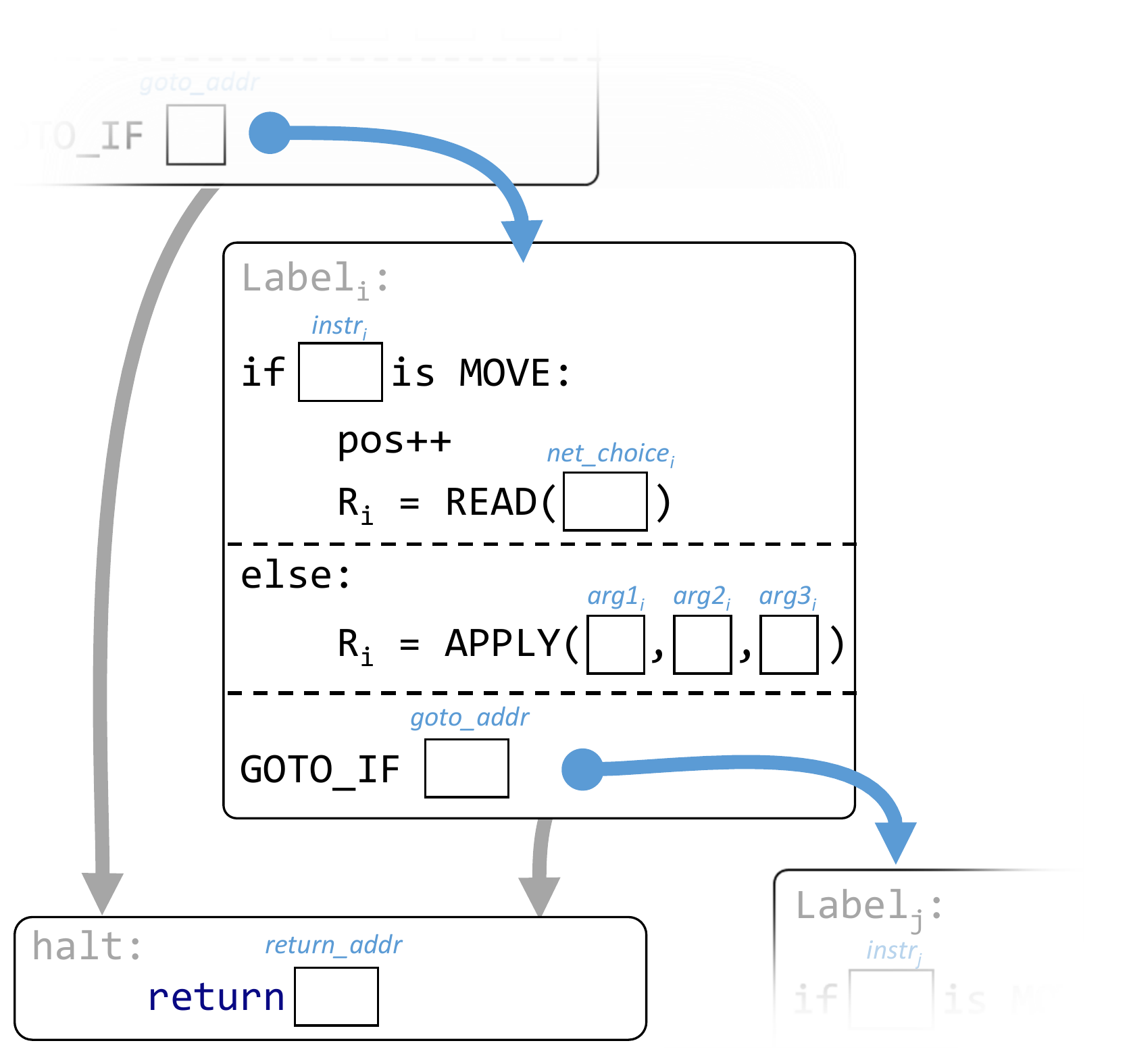}
\end{subfigure}
\begin{subfigure}[t]{.17\textwidth}
(b) \\
\includegraphics[width=1.2in]{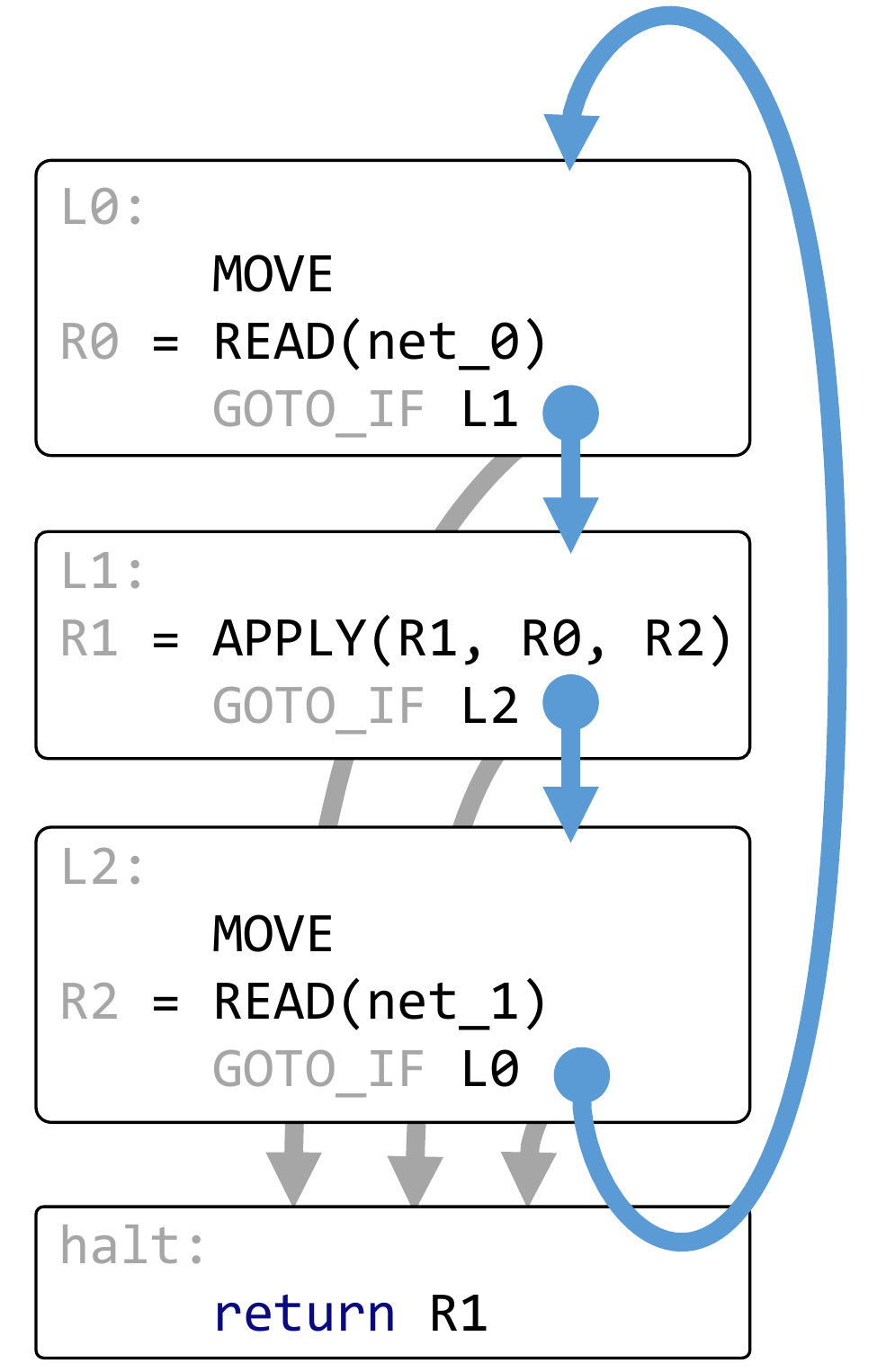}
\end{subfigure}
\caption{\small Overview of the \mathScenario model. (a) The general form of a block in the model. Blue elements are learnable. (b) A loop-based solution to the task in the \mathScenario scenario.\label{fig:MATHSsolutions}}
\vspace{-3ex}
\end{figure}

The final task investigates the synthesis of more complex, loopy control flow. A natural solution to execute the expression on the tape is to build a loop with a body that alternates between moving the head and applying the operators [see \figref{fig:MATHSsolutions}(b)]. This loopy solution has the advantage that it generalizes to handle arbitrary length arithmetic expressions.

\figref{fig:MATHSsolutions}(a) shows the basic architecture of the interpreter used in this scenario. We provide a set of blocks each containing the instruction \texttt{MOVE} or \texttt{APPLY}, an address, a register and a \texttt{net\_choice}. A \texttt{MOVE} instruction increments the position of the head and loads the new symbol into a block's register using either \texttt{net\_0} or \texttt{net\_1} as determined by the block's \texttt{net\_choice}. After executing the instruction, the interpreter executes a \texttt{GOTO\_IF} statement which checks whether the head is over the end of the tape and if not then it passes control to the block specified by \texttt{goto\_addr}, otherwise control passes to a \texttt{halt} block which returns a chosen register value and exits the program. This model describes a space of $\sim 10^6$ syntactically distinct programs.

\section{Baselines}
\label{sec:baselines}
To evaluate the merits of including the source code structure in \neuralterpretshort models, we build baselines that replace the differentiable program interpreter with neural networks, thereby creating purely neural solutions to the lifelong PPBE tasks. As in the \neuralterpretshort case, we specialize the neural baselines for either the $2 \times 2$ tasks (with emphasis on lifelong learning) or the \mathScenario task (with emphasis on generalization).

\subsection{$2 \times 2$ baselines}
We define a \textit{column} as the following neural architecture (see \figref{fig:baselines}(a)):
\begin{itemize}
\item Each of the images in the $2\times 2$ grid is passed through an embedding network with 2 layers of 256 neurons (cf. \texttt{net\_0/1}) to produce a 10-dimensional embedding. The weights of the embedding network are shared across all 4 images.
\item These 4 embeddings are concatenated into a 40-dimensional vector and for the \arithScenario the auxiliary integers are represented as one-hot vectors and concatenated with this 40-dimensional vector.
\item This is then passed through a network consisting of 3 hidden layers of 128 neurons to produce a 19-dimensional output.
\end{itemize}
We construct 3 different neural baselines derived from this column architecture (see \figref{fig:baselines}):

\begin{figure*}
\centering
\includegraphics[width=5in]{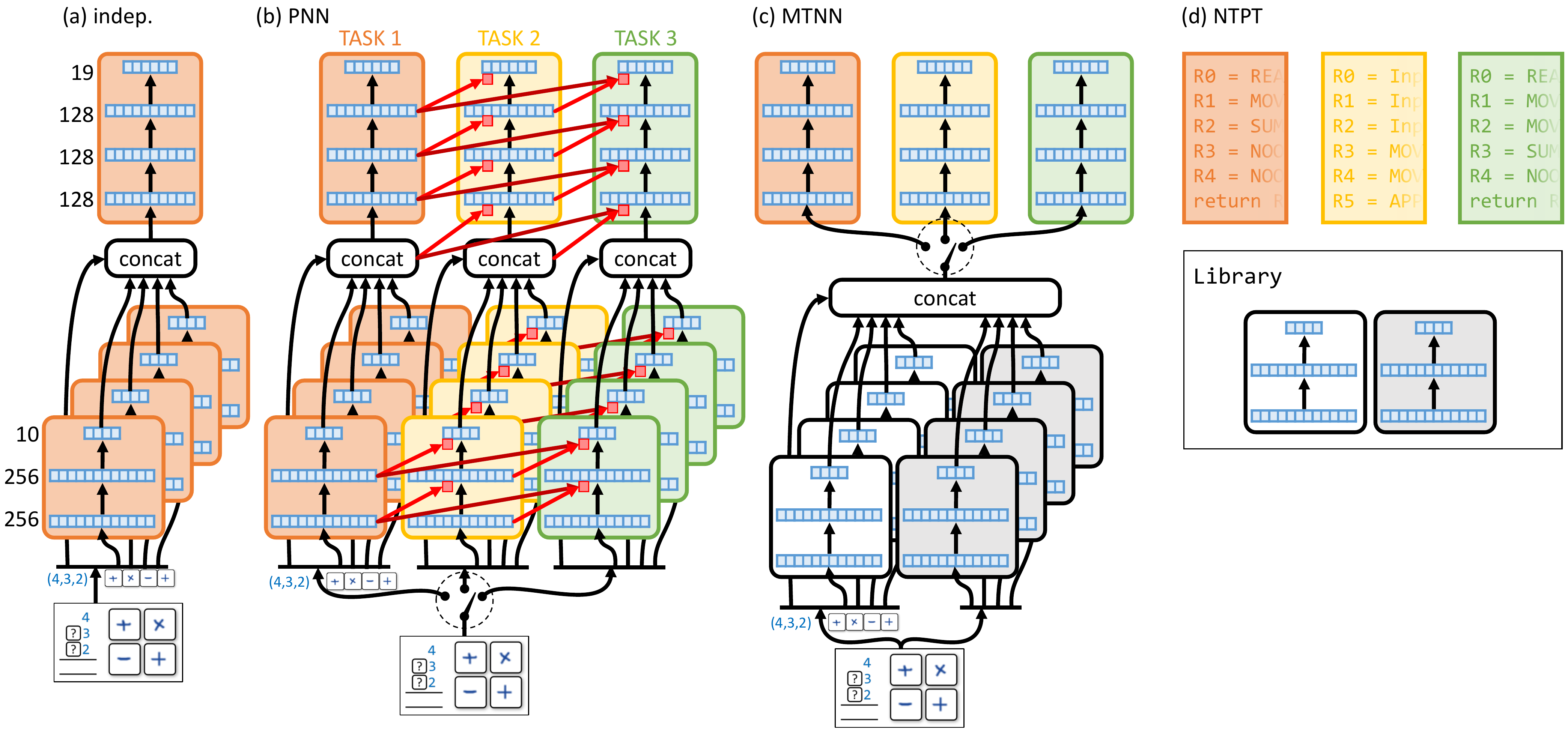}
\caption{\small Cartoon illustration of all models used in the $2\times 2$ experiments. See text for details.\label{fig:baselines}}
\end{figure*}

\begin{enumerate}
\item \textbf{Indep.}: Each task is handled by an independent column with no mechanism for transfer.
\item \textbf{Progressive Neural Network (PNN)}: We follow \cite{rusu16} and build lateral connections linking each task specific column to columns from tasks appearing earlier in the learning lifetime. Weights in all columns except the active task's column are frozen during a training update. Note that the number of layers in each column must be identical to allow lateral connections, meaning we cannot tune the architecture separately for each task.
\item \textbf{Multitask neural network (MTNN)}: We split the column into a shared perceptual part and a task specific part. The perceptual part consists of \texttt{net\_0} and \texttt{net\_1} embedding networks (note that we use a similar symmetry breaking technique mentioned in \secref{sec:shared} to encourage specialization of these networks to either digit or operator recognition respectively).

The task-specific part consists of a neural network that maps the perceptual embeddings to a 19 dimensional output. Note that unlike PNNs, the precise architecture of the task specific part of the MTNN can be tuned for each individual task. We consider two MTNN architectures:
\begin{enumerate}
\item MTNN-1: All task-specific parts are 3 layer networks comparable to the PNN case.
\item MTNN-2: We manually tune the number of layers for each task and find best performance when the task specific part contains 1 hidden layer for the \sumsScenario tasks and 3 layers for the \arithScenario tasks.
\end{enumerate}
\end{enumerate}

\subsection{\mathScenario baselines}

For the \mathScenario task, we build purely neural baselines which (1) have previously been shown to offer competitive generalization performance for some tasks with sequential inputs of varying length (2) are able to learn to execute arithmetic operations and (3) are easily integrated with the library of perceptual networks learned in the $2 \times 2$ tasks. We consider two models fulfilling these criteria: an LSTM and a Neural GPU.

For the LSTM, at each image in the mathematical expression the network takes in the embeddings of the current symbol from \texttt{net\_0} and \texttt{net\_1}, updates an LSTM hidden state and then proceeds to the next symbol. We make a classification of the final answer using the last hidden state of the LSTM. Our best performance is achieved with a 3 layer LSTM with 1024 elements in each hidden state and dropout between layers. 

For the Neural GPU, we use the implementation from the original authors\footnote{available at \url{https://github.com/tensorflow/models/tree/master/neural_gpu}} \cite{Kaiser15}.

\section{Experiments}
\label{sec:experiments}
In this section we report results illustrating the key benefits of \neuralterpretshort for the lifelong PPBE tasks in terms of knowledge transfer (\secref{sec:lifelong}) and generalization (\secref{sec:generalization}).

\begin{figure*}
\vspace{-2mm}
\begin{tabular}{m{4in}|m{1in}}
\centering\includegraphics[width=4in]{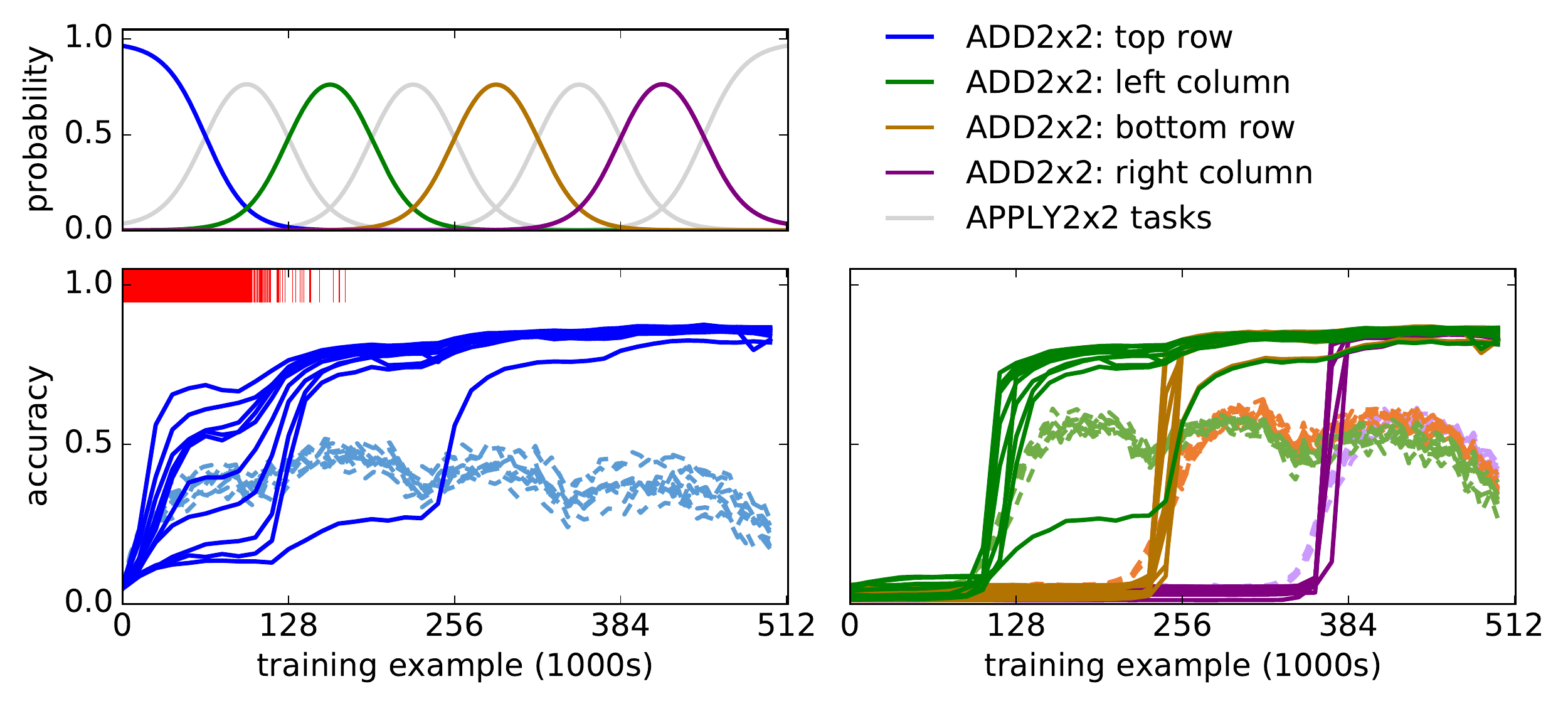}&
\centering\includegraphics[width=2in]{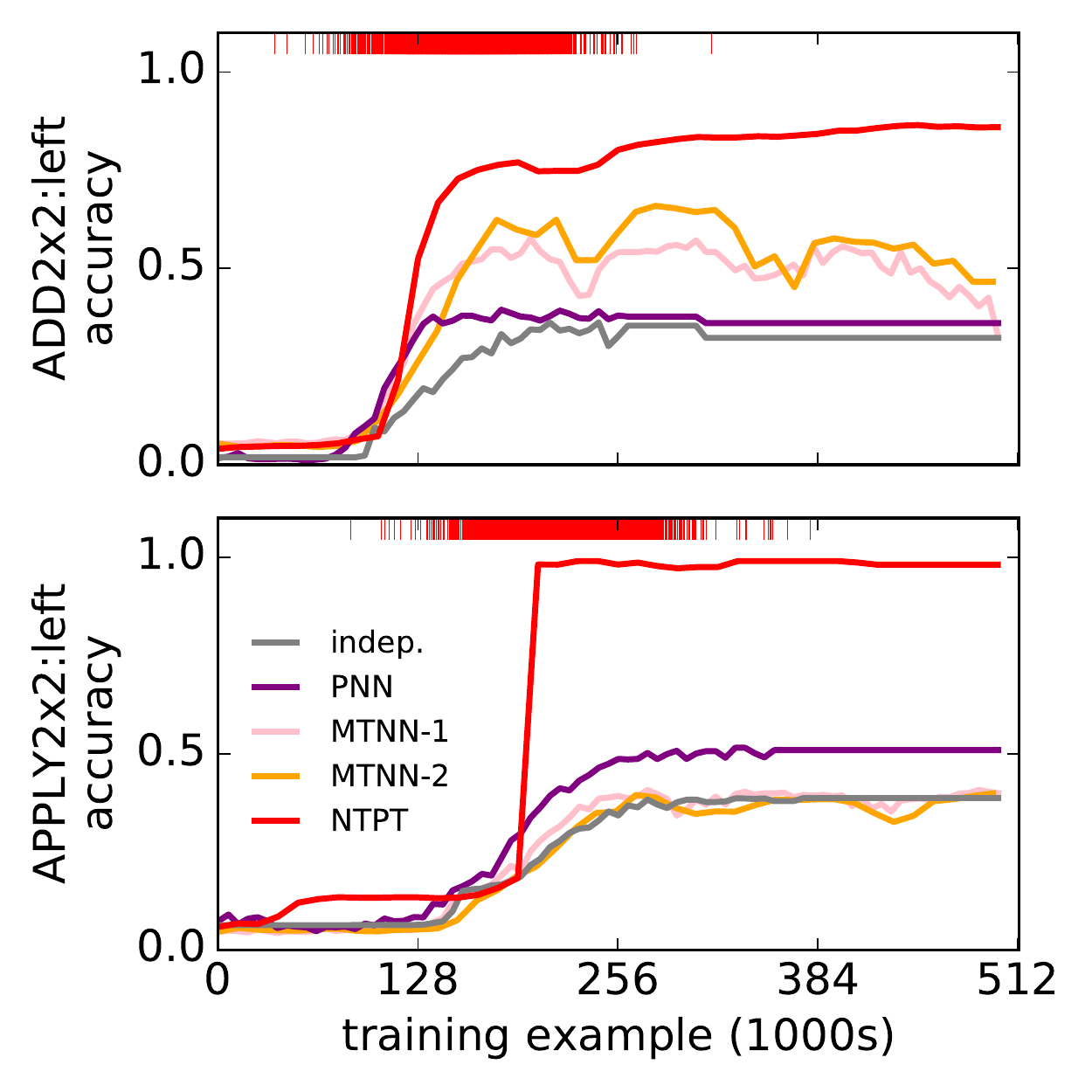}
%
\end{tabular}

\hspace{0.65in} (a) \hspace{1.65in} (b) \hspace{1.9in} (c)
\caption{\small Lifelong learning with \neuralterpretshort. (a) top: the sequential learning schedule for all 8 tasks, bottom: performance of \neuralterpretshort (solid) and the MTNN-2 baseline (dashed) on the first \sumsScenario task. (b) performance on the remaining \sumsScenario tasks. (c) Performance of all the baselines on the *:left tasks.\label{fig:lifelongResults}}
\vspace{-5mm}
\end{figure*}

\subsection{Lifelong Learning}
\label{sec:lifelong}

Demonstration of lifelong learning requires a series of tasks for which there is insufficient data to learn independent solutions to all tasks and instead, success requires transferring knowledge from one task to the next. Empirically, we find that training any of the purely neural baselines or the \neuralterpretshort model on individual tasks from the \sumsScenario scenario with only 1k distinct $2\times2$ examples produces low accuracies of around $40\pm20\%$ (measured on a held-out test set of 10k examples). Since none of our models can satisfactorily solve an \sumsScenario task independently in this small data regime, we can say that any success on these tasks during a lifetime of learning can be attributed to successful knowledge transfer. In addition, we check that in a data rich regime (e.g. $\geq$4k examples) all of the baseline models and \neuralterpretshort can independently solve each task with $>$80\% accuracy. This indicates that the models all have sufficient capacity to represent satisfactory solutions, and the challenge is to find these solutions during training.

We train on batches of data drawn from a time-evolving probability distribution over all 8 tasks in the $2 \times 2$ scenarios (see the top of \figref{fig:lifelongResults}(a)). During training, we observe the following key properties of the knowledge transfer achieved by \neuralterpretshort:

\textbf{Reverse transfer}: \figref{fig:lifelongResults}(a) focuses on the performance of \neuralterpretshort on the first task (\sumsScenario:top). The red bars indicate times where the the system was presented with an example from this task. Note that even when we have stopped presenting examples, the performance on this task continues to increase as we train on later tasks - an example of \textit{reverse transfer}. We verify that this is due to continuous improvement of \texttt{net\_0} in later tasks by observing that the accuracy on the \sumsScenario:top task closely tracks measurements of the accuracy of \texttt{net\_0} directly on the digit classification task.

\begin{figure}[h!]
  \centering
\tiny
\vspace{0ex}
\setlength\tabcolsep{0.1cm}
  \begin{tabular}{ccccccc}
\toprule
& task & indep & PNN & MTNN-1 & MTNN-2 & NTPT \\
\midrule
\multirow{4}{*}{\rotatebox{90}{\sumsScenario}} & top &  35\% & 35\% & 26\% & 24\% & 87\%\\
& left & 32\% & 36\% & 38\% & 47\% & 87\%\\
& bottom & 34\% & 33\% & 40\% & 56\% & 86\%\\
& right & 32\% & 35\% & 44\% & 60\% & 86\%\\
\midrule
\multirow{4}{*}{\rotatebox{90}{\arithScenario}} & top & 38\% & 39\% & 40\% & 38\% & 98\%\\
& left & 39\% & 51\% & 41\% & 39\% & 100\%\\
& bottom & 39\% & 48\% & 41\% & 40\% &100\%\\
& right & 39\% & 51\% & 42\% & 37\% & 100\%\\
\bottomrule
\end{tabular}
  \captionof{figure}{\small Final accuracies on all $2\times 2$ tasks for all models at the end of lifelong learning}
  \label{fig:accuracies}
\vspace{-3mm}
\end{figure}
\textbf{Avoidance of catastrophic forgetting}: \figref{fig:lifelongResults}(b) shows the performance of the \neuralterpretshort on the remaining \sumsScenario tasks. Both \figref{fig:lifelongResults}(a) and (b) include results for the MTNN-2 baseline (the best baseline for these tasks). Note that whenever the dominant training task swaps from an \sumsScenario task to an \arithScenario task the baseline's performance on \sumsScenario tasks drops. This is because the shared perceptual network becomes corrupted by the change in task - an example of \textit{catastrophic forgetting}. To try to limit the extent of catastrophic forgetting and make the shared components more robust, we have a separate learning rate for the perceptual networks in both the MTNN baseline and \neuralterpretshort which is 100 fold smaller than the learning rate for the task-specific parts. With this balance of learning rates we find empirically that \neuralterpretshort does not display catastrophic forgetting, while the MTNN does.

\textbf{Final performance}: \figref{fig:lifelongResults}(c) focuses on the \sumsScenario:left and \arithScenario:left tasks to illustrate the relative performance of all the baselines described in \secref{sec:baselines}. Note that although PNNs are effective at avoiding catastrophic forgetting, there is no clear overall winner between the MTNN and PNN baselines. \neuralterpretshort learns faster and to a higher accuracy than all baselines for all the tasks considered here. For clarity we only plot results for the *:left tasks: the other tasks show similar behavior and the accuracies for all tasks at the end of the lifetime of learning are presented in \figref{fig:accuracies}.

\subsection{Generalization}
\label{sec:generalization}
\begin{figure}
\centering
\includegraphics[width=2.2in]{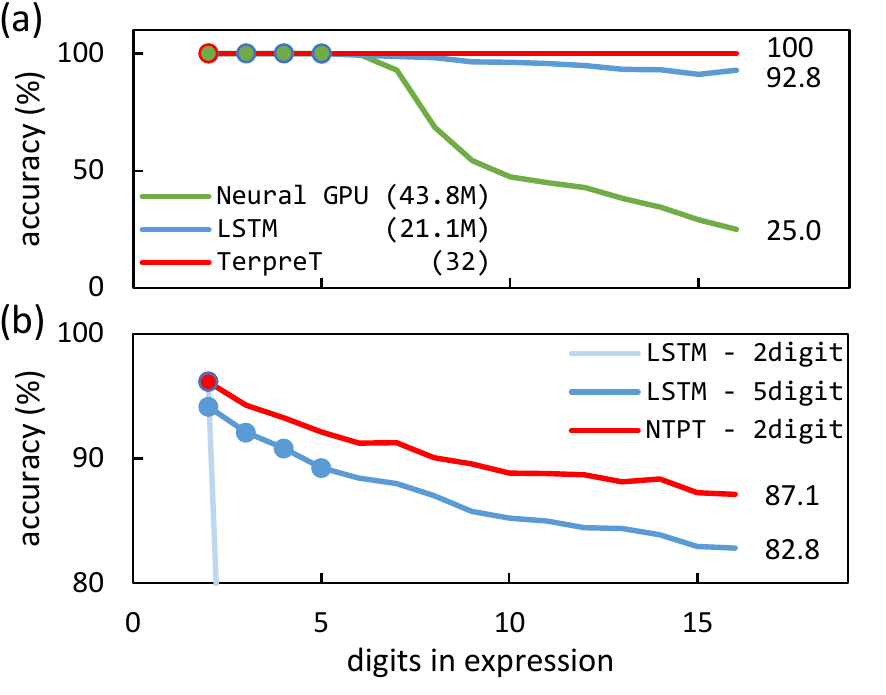}
\vspace{-2mm}
  \captionof{figure}{\small Generalization behavior on \mathScenario expressions. Solid dots indicate expression lengths used in training. We show results on (a) a simpler non-perceptual \mathScenario task (numbers in parentheses indicate parameter count in each model) and (b) the \mathScenario task including perception.}
  \label{fig:generalizationResults}
\vspace{-5mm}
\end{figure}
In the final experiment we take \texttt{net\_0/1} from the end of the \neuralterpretshort $2\times 2$ training and start training on the \mathScenario scenario. For the \neuralterpretshort model we train on arithmetic expressions containing only 2 digits. The known difficulty in training differentiable interpreters with free loop structure \cite{Gaunt16} is revealed by the fact that only $2/100$ random restarts converge on a correct program in a global optimum of the loss landscape. We detect convergence by a rapid increase in the accuracy on a validation set (typically occurring after around 30k training examples). Once the correct program is found, continuing to train the model mainly leads to further improvement in the accuracy of \texttt{net\_0}, which saturates at 97.5\% on the digit classification task. The learned source code provably generalizes perfectly to expressions containing any number of digits, and the only limitation on the performance on long expressions comes from the repeated application of the imperfect \texttt{net\_0}.

To pick a strong baseline for the MATH problem, we first perform a preliminary
experiment with two simplifications: (1) rather than expecting strong
generalization from just 2-digit training examples, we train candidate baselines
with supervision on examples of up to 5 digits and 4 operators, and (2) we
remove the perceptual component of the task, presenting the digits and operators
as one-hot vectors rather than images.
\figref{fig:generalizationResults}(a) shows the generalization performance of
the LSTM and Neural GPU (512-filter) baselines in this simpler setting after
training to convergence.\footnote{Note that \cite{Price16} also find poor
  generalization performance for a Neural GPU applied to the similar task of
  evaluating arithmetic expressions involving binary numbers.}
Based on these results, we restrict attention to the LSTM baseline and return to
the full task including the perceptual component.
In the full \mathScenario task, we initialize the embedding networks of each model
using \texttt{net\_0/1} from the end of the \neuralterpretshort $2\times 2$
training.
\figref{fig:generalizationResults}(b) shows generalization of the
\neuralterpretshort and LSTM models on expressions of up to 16 digits (31
symbols) after training to convergence.
We find that even though the LSTM shows surprisingly effective generalization
when supplied supervision for up to 5 digits, \neuralterpretshort trained on
only 2-digit expressions still offers better results.




\section{Related work}

\paragraph{Lifelong Machine Learning. }
%
We operate in the paradigm of Lifelong Machine Learning (LML)
\citep{Thrun94,Thrun95,Thrun96,Silver13,Chen15}, where a learner is presented a sequence
of different tasks and the aim is to retain and re-use knowledge from earlier tasks to more efficiently
and effectively learn new tasks. This is distinct from related paradigms of
multitask learning (where a set of tasks is presented rather than in sequence \citep{Caruana97, kumar12, luong2015multi, rusu16}), transfer learning (transfer of knowledge from a source to target domain without notion of knowledge retention \citep{pan10}), and curriculum learning (training a single model for a single task of varying difficulty \citep{Bengio09}). 

The challenge for LML with neural networks is the problem of catastrophic forgetting: if the distribution of examples changes during training, then neural networks are prone to forget knowledge gathered from early examples.
Solutions to this problem involve instantiating a knowledge repository (KR) either directly storing data from earlier tasks or storing (sub)networks trained on the earlier tasks with their weights frozen. This knowledge base allows either (1) rehearsal on historical examples \citep{robins95}, (2) rehearsal on virtual examples generated by the frozen networks \citep{silver02, silver06} or (3) creation of new networks containing frozen sub networks from the historical tasks \citep{rusu16, shultz01}

To frame our approach in these terms, our KR contains partially-trained neural network classifiers which we call from learned source code. Crucially, we never freeze the weights of the networks in the KR: all parts of the KR can be updated during the training of all tasks - this allows us to improve performance on earlier tasks by continuing training on later tasks (so-called reverse transfer). Reverse transfer has been demonstrated previously in systems which assume that each task can be solved by a model parameterized by an (uninterpretable) task-specific linear combination of shared basis weights \citep{ruvolo13}. The representation of task-specific knowledge as source code, learning from weak supervision, and shared knowledge as a deep neural networks distinguishes this work from the linear model used in \cite{ruvolo13}.

\comment{
either solve a multitask problem at each step that learns from the current task and all historical tasks
at every step [GO-MTL, (Kumar et al., 2012)], or to fix what was learned in previous tasks [ELLA, Progressive Networks].

LML is often realized by keeping a knowledge base (KB) which is updated after
each task is learned, and whose contents can be accessed when learning later
tasks.
In the simplest case, the KB simply stores all previous examples, in which case
multitask learning can be applied to the historical data. Alternatively, the KB can store   \cite{}

In other cases, logical relations [NELL?].
To frame our approach in these terms, the KB would contain partially-trained neural network classifiers.
[TODO: cite some other specifics].




The key differences in our approach:
\begin{itemize}
\item We operate in a LML setting rather than a repeated multitask setting; that is, we do not store data in the KB.
\item All parts of the KB can be updated during the training of all tasks. (This distinguishes us from Progressive Networks? and it allows us to improve performance on earlier tasks by continuing training on later tasks)
\item While LML has been considered in the context of many types of supervision (unsupervised, semi-supervised, RL), we are not aware of work that addresses PBE in the LML context.
\end{itemize}
}

\vspace{-0.3cm}
\paragraph{Neural Networks Learning Algorithms.}
Recently, extensions of neural networks with primitives such as memory and
discrete computation units have been studied to learn algorithms from
input-output data~\citep{Graves14,Weston14,Joulin15,Grefenstette15,Kurach15,Kaiser15,Reed15,Bunel16,Andrychowicz16,Zaremba16,Graves16,Riedel16,Gaunt16,Feser16}.
A dominant trend in these works is to use a neural network controller to managing differentiable computer architecture. We flip this relationship, and in our approach, a differentiable interpreter acts as the controller that can make calls to neural network components.

The methods above, with the exception of \cite{Reed15} and \cite{Graves16},
operate on inputs of (arrays of) integers.
However, \cite{Reed15} requires extremely strong supervision,
where the learner is shown all intermediate steps to solving a problem; our learner only observes input-output examples. \cite{Reed15} also show the performance of their system in a multitask setting. In some cases, additional tasks harm performance of their model and they freeze parts of their model when adding to their library of functions. 
Only \cite{Bunel16}, \cite{Riedel16} and \cite{Gaunt16} aim to consume and produce source code that
can be provided by a human (e.g. as sketch of a solution) to or returned to a human (to potentially provide feedback).

\vspace{-2mm}
\section{Discussion}
We have presented \neuralterpret, a framework for building end-to-end trainable models that structure their solution as a source code description of an algorithm which may make calls into a library of neural functions.
Experimental results show that these models can successfully be trained in a lifelong learning context, and they are resistant to catastrophic forgetting; in fact, they show that even after instances of earlier tasks are no longer presented to the model, performance still continues to improve.

\rev{
Our experiments concentrated on two key benefits of the hybrid representation of task solutions as source code and neural networks. First, the source code structure imposes modularity which can be seen as \emph{focusing the supervision}.
If a component is not needed for a given task, then the differentiable interpreter can choose not to use it, which shuts off any gradients from flowing to that component.
We speculate that this could be a reason for the models being resistant to catastrophic forgetting, as the model either chooses to use a classifier, or ignores it (which leaves the component unchanged). The second benefit is that learning programs imposes a bias that favors learning models that exhibit strong generalization. Additionally, the source code representation has the advantage of being interpretable by humans, allowing verification and incorporation of domain knowledge describing the shape of the problem through the source code structure.
}

\rev{
The primary limitation of this design is that it is known that differentiable interpreters are difficult to train on problems significantly more complex than those presented here \citep{Kurach15,neelakantan15,Gaunt16}.
However, if progress can be made on more robust training of differentiable interpreters (perhaps extending ideas in \cite{neelakantan15} and \cite{Feser16}),
then we believe there to be great promise in using hybrid models to build large lifelong learning systems.
}



\bibliography{references}
\bibliographystyle{icml2017}

\end{document}